\documentclass{evolang12}
\usepackage{graphicx}
\usepackage{url}
\usepackage{authblk}
\usepackage{amsthm}
\usepackage{adjustbox}
\usepackage{multirow}
\usepackage{multicol}
\usepackage{arydshln}

\begin{document}

\title{Simulating Lexical Semantic Change from Sense-Annotated Data}
\author[*]{DOMINIK SCHLECHTWEG}
\author[ ]{SABINE SCHULTE IM WALDE}
\affil[*]{dominik.schlechtweg@ims.uni-stuttgart.de}
\affil[ ]{Institute for Natural Language Processing, University of Stuttgart, Germany}

\maketitle

\abstracts{We present a novel procedure to simulate lexical semantic change from synchronic sense-annotated data, and demonstrate its usefulness for assessing lexical semantic change detection models. The induced dataset represents a stronger correspondence to empirically observed lexical semantic change than previous synthetic datasets, because it exploits the intimate relationship between synchronic polysemy and diachronic change. We publish the data and provide the first large-scale evaluation gold standard for LSC detection models.}

\section{Introduction}
\label{sec:intro}

Evaluating Lexical Semantic Change (LSC) detection models is notoriously challenging. Existing testsets are flawed because they are too small to allow for generalizing over the results obtained on them. Artificial data, on the other hand, can be created in larger quantities, but typically relies on assumptions that may or may not be correct, such as the strength of semantic relatedness that old and new senses in LSC have. A clear advantage of artificial data is, however, that it allows the precise control of potentially influencing variables such as frequency and polysemy. 

After spelling out the implicit assumptions of previous work, this paper presents a novel procedure to simulate lexical semantic change from synchronic sense-annotated data, which we consider more realistic than in earlier approaches. By splitting the synchronic data into two parts reflecting different sense frequency distributions for a word we simulate sense divergences. In a second stage, we define a graded and a binary notion of LSC based on differences between the obtained sense frequency distributions. These notions are then used to calculate the gold scores determining for each sense-annotated word the degree of change and whether senses were gained or lost. With the proposed definitions, we hope to provide a solid foundation for the basic concepts in the field of LSC detection.

\section{Related Work}
\label{sec:previous}

Most previous evaluations for LSC detection models rely on small amounts of empirically observed data, which was either hand-selected \shortcite{Sagi09p104,Jatowt:2014,Hamilton16,Hamilton:2016,Frermann:2016,del2017semantic} or annotated by humans \shortcite{Cook14p1624,schlechtweg-EtAl:2017:CoNLL,Tahmasebi17,Schlechtwegetal18,Perrone19}. An alternative approach is synthetic evaluation, where pseudo-change is simulated by collapsing uses of different words \shortcite{Cook10,Kulkarni14,RosenfeldE18,Dubossarskyetal19,Shoemark2019}. This procedure is very similar to the creation of pseudo-polysemy in word sense disambiguation \shortcite{Schutze1998,pilehvar2013paving}:
Changes in the senses a word $w$ expresses are artificially created by copying a corpus and relabeling the uses of another word $w'$ (known to express different senses) as uses of $w$ in the copy. The word $w$ is then guaranteed to express a different sense in the copy and the difference between the two corpora can be equated with artificial LSC.\footnote{\shortciteA{RosenfeldE18} use a slightly different variation of this procedure.} 
This procedure mimics the changes in word senses occurring in empirical LSC, but requires assumptions about other factors. One such factor is the choice of words that should be collapsed and what their semantic relation should be. With the exception of \shortciteA{Dubossarskyetal19}, all of the existing work collapses uses of words that have no semantic relation. This has the advantage that strong differences in the contextual distribution of words are created, which are more easily detected by computational models. However, there is plenty of evidence showing that LSC does not introduce random new senses of a word, but that new senses are very often semantically related to one of the old senses \shortcite{Blank97XVI}. Hence, the changes to the contextual distribution of a word in empirical LSC are often more subtle and hard to detect than the ones introduced in previous synthetic evaluations.

\section{Simulation}
\label{sec:simulation}

Polysemy is the synchronic result of lexical semantic change \shortcite{Blank97XVI,Bybee:2015aa}. Accordingly, the different senses a word may express today have been developed some time in the past by the word undergoing a process of LSC. We exploit this idea by using the modern (synchronic) senses of a polysemous word to simulate LSC, i.e., we reconstruct the diachronic process using its synchronic result. In this way, we (i) guarantee that the different senses used in the simulation are likely to be semantically related, as the different senses of a word are usually semantically related to each other \shortcite{Fillmore2000}. And (ii) we simulate divergences for senses which have empirically been attached to the same word and are thus probable candidates to occur in empirical LSC.

\subsection{Corpus}
\label{sec:corpus}

To simulate LSC in the above-described way we need sense-annotated data. We use \textbf{SemCor}, a sense-tagged corpus of English \shortcite{Langoneetal04} which represents a subset of the Brown Corpus \shortcite{francis79browncorpus}. SemCor contains 700,000 tokens, of which more than 200,000 are sense-annotated; the corpus is lemmatized and POS-tagged. Similar corpora in other languages \shortcite[e.g.]{bentivogli2005,Henrich2013} would allow to easily extend our methodology to create evaluation data for further languages.

\subsection{Two concepts of LSC}
\label{sec:concepts}

\paragraph{Graded.} In order to simulate LSC, we first need a definition of what we mean by LSC. Previous research has not been explicit about the underlying concepts and often relied on an intuitive notion of \textit{degree of LSC} \shortcite{Hamilton:2016,dubossarsky2017,bamler17,RudolphB18,RosenfeldE18,Schlechtwegetal18}. A well-defined concept of graded LSC, however, should enable us to compare any two words over time and decide which of the words changed more. Hence, it should provide an answer to questions like: Did a word that lost a very frequent sense change more than a word that lost a very infrequent sense? And did a word that gained two senses change less than a word that lost three senses? The field is still lacking such a definition of the degree of LSC of a word.

\paragraph{Binary.} This graded notion of LSC seems to diverge from the definition applied in historical linguistics, where LSC is typically not assumed to be graded, but binary \shortcite[cf.~p.~113]{Blank97XVI}. That is, either a word gained/lost a sense over time, or not, while in computational linguistics slight changes to the frequencies of different word senses are usually also considered as instances of LSC (hence the term \textit{degree of LSC}).\footnote{Note for completeness that some work in computational linguistics also assumes a binary notion \shortcite{Cook14p1624,Tahmasebi17,Perrone19,Shoemark2019}.} This deviation is striking, as the most straightforward application of LSC detection models is their use to aid historical linguists \shortcite{Hamilton:2016}.
Yet, the graded notion is applicable to related tasks, such as detecting the early stages of a meaning loss (as indicated by sense frequency decrease).

\begin{table}[htp]
  \tablecaption{Corpus sample for the noun \textit{plant}.}
  {
\begin{adjustbox}{width=1.0\linewidth}
\begin{tabular}{@{}l@{}}\hline
 This reduces the number of expensive \textbf{plant} shutdowns and startups. \hfill \textbf{(s1)} \\
 The pilot \textbf{plant} was equipped with a 3-hp. turbine aerator (Figure 2). \hfill \textbf{(s1)} \\
 Remove about half the branches from each \textbf{plant}, leaving only the strongest with the largest buds. \hfill \textbf{(s2)} \\
 ``On the side toward the horizon -- the southern hemisphere -- it is spring; \textbf{plants} are being taught to grow''. \hfill \textbf{(s2)}\\
 Can you share medical facilities and staff with neighboring \textbf{plants}?? \hfill \textbf{(s1)} \\\hline
\end{tabular}\label{tab:plant0}
\end{adjustbox}}
\end{table}%

\subsubsection{Sense Frequency Distributions}

We propose quantitative definitions of the two notions described above. The definitions are based on the concept of a Sense Frequency Distribution (SFD) \shortcite[e.g.]{mccarthy2004,lau-EtAl:2014}. A SFD encodes how often a word $w$ occurs in each of its senses. Consider the word \textit{plant}. In SemCor it occurs expressing either of two senses:
\begin{itemize}
\item \textbf{sense 1}: plant, works, industrial plant (buildings for carrying on industrial labor); ``they built a large plant to manufacture automobiles''\footnote{\url{https://wordnet.princeton.edu/}}
\item \textbf{sense 2}: plant, flora, plant life (botany: a living organism lacking the power of locomotion) 
\end{itemize}
This means that each use of \textit{plant} in the corpus is assigned to one of these two senses, as we can see in the sample in Table \ref{tab:plant0}: \textit{plant} occurs three times in sense 1 and two times in sense 2. Hence, its SFD is provided as (3,2). Generally, given a set of $w$'s uses $U$ where each use is assigned to one of the senses in the sequence $S=(S_1,S_2,...,S_i$), $w$'s SFD is defined by
\begin{equation*}
T = (f(S_1),f(S_2),...,f(S_i))
\end{equation*}
where $f(S_i)$ is the number of times any use from $U$ was mapped to the $i$th sense in $S$.

\paragraph{Graded LSC.} For two different sets of uses $U_1$ and $U_2$ the corresponding SFDs may differ. We will now define two measures quantifying the difference between any two same-sized SFDs, which will correspond to graded and binary LSC. Assume we have two SFDs $T_1$ and $T_2$ defined for the same word $w$ and sense sequence $S$, but for two \textit{different} use sets $U_1$ and $U_2$. We first normalize $T_1$ and $T_2$ to probability distributions $P$ and $Q$ by dividing each element by the total sum of the frequencies of all senses in the respective distribution. The degree of LSC of the word $w$ is then defined as the Jensen-Shannon distance between the two normalized frequency distributions:
\begin{equation*}
G(w)=JSD(P,Q)
\end{equation*}
where the Jensen-Shannon distance is the symmetrized square root of the Kullback-Leibler divergence \shortcite{Lin1991,DonosoS17}.\footnote{We prefer the Jensen-Shannon distance over Kullback-Leibler divergence, because the former is a true metric in contrast to the latter. This means that amongst other properties specific to metrics the Jensen-Shannon distance is symmetric, i.e., $JSD(X,Y)=JSD(Y,X)$.}
$G(w)$ ranges between $0$ and $1$ and is high if $P$ and $Q$ assign very different probabilities to the same senses.

\paragraph{Binary LSC.} Binary LSC of the word $w$ is then defined as 
\begin{equation*}
  \begin{split}
  B(w)= 1 & \textnormal{ if for some $i$, ${P}_i = 0.0$ and ${Q}_i \geq k$,}\\
    & \textnormal{or vice versa.}\\
  B(w)= 0 & \textnormal{ else.}
  \end{split}
\end{equation*}
where $X_i$ is the $i$th element in $X$ and $k$ is a probability threshold set to $0.1$. That is, $B(w)$ will be $1$ if there is a sense which has at least probability of $0.1$ in either $P$ or $Q$ but a probability of $0.0$ in the other (a meaning is gained or lost). If this is not the case, $B(w)$ will be $0$. Note that $B(w)$ can be seen as an extreme special case of $G(w)$: if a sense is never assigned to any use in $U_1$, this sense will have probability $0.0$ in $P$, which will cause $G(w)$ to be higher than for any other possible probability assignment to that sense. However, $G(w)$ will typically not be equal to $1.0$ in such a case, as it is also sensitive to the probabilities of the other senses which may have changed only slightly or not at all.

\subsection{Algorithm}
\label{sec:algorithm}

In order to simulate LSC in SemCor we split it into two parts ($C_1$, $C_2$) and extract the corresponding SFDs ($T_1$, $T_2$) for each sense-annotated lemma $w$. From $T_1$ and $T_2$ the scores of above-defined gold notions of LSC follow directly. The splitting process has two steps: In step (i), we introduce strong changes for specific target lemmas. For this we sample all lemmas with a frequency between 100 and 1000 and split their sentences into senses, i.e., for each target lemma we randomly shuffle senses and split them at a random index into two subsets. Then we try to assign sentences with senses from the first subset to $C_1$ and sentences with senses from the second subset to $C_2$. This maximizes change, because senses will tend to have uses in only one of $C_1$, $C_2$.
In step (ii), all remaining sentences are randomly shuffled, split in half and added to $C_1$, $C_2$ respectively. With this process a non-target lemma will tend to have a rather low change score.

Finally, we extract $C_1$ and $C_2$ sentence-wise to separate text files, and the LSC scores to a CSV file. For each sentence, all words (no punctuation) are extracted and replaced by their lemma if existent, else the lowercased token is extracted. Annotated phrases like \textit{on\_the\_other\_hand} are split into individual words to increase data size, i.e., \textit{on the other hand}. The resulting corpora $C_1$ and $C_2$ have sizes with $0.34$M and $0.36$M tokens.

Consider Tables \ref{tab:split} and \ref{tab:split2}, showing the sample corpus from Table \ref{tab:plant0} split according to steps (i) and (ii) respectively. In Table \ref{tab:split}, \textit{plant} is treated as a target lemma and split according to step (i): sense 1 is assigned to $C_2$, while sense 2 is assigned to $C_1$. The resulting SFDs are $T_1=(0,2)$ and $T_2=(3,0)$, creating high change scores of $G(\textnormal{\textit{plant}})=1.0$ and $B(\textnormal{\textit{plant}})=1$. As the probability changes of the two senses (from $0.0$ to $1.0$ and vice versa) are the strongest possible, the graded change score is at its maximum. And as \textit{plant} also loses and gains a sense from $C_1$ to $C_2$ it shows binary change.

\begin{table}[htp]
  \tablecaption{Sample corpus split for the target lemma \textit{plant}. $T_1=(0,2)$, $T_2=(3,0)$, $G(w)=1.0$ and $B(w)=1$.}
  {
\begin{adjustbox}{width=1.0\linewidth}
\begin{tabular}{@{}p{6.8cm}p{6.8cm}@{}}\hline
 \centering $C_1$ & \centering $C_2$ \tabularnewline\hline
remove about half the branch from each \textbf{plant} leave only the strong with the largest bud \hfill \textbf{(s2)} & the pilot \textbf{plant} was equip with a 3 hp turbine aerator figure 2 \hfill \textbf{(s1)} \\
on the side toward the horizon the southern hemisphere it be spring \textbf{plant} are being teach to grow \hfill \textbf{(s2)} & this reduce the number of expensive \textbf{plant} shutdown and startup \hfill \textbf{(s1)}	\\
 & can you share medical facility and staff with neighboring \textbf{plant} \hfill \textbf{(s1)} \\\hline
\end{tabular}\label{tab:split}
\end{adjustbox}}
\end{table}%

\noindent In Table \ref{tab:split2}, \textit{plant} is treated as a non-target lemma and thus split according to step (ii): both senses are assigned uniformly to $C_1$, $C_2$. The resulting SFDs are $T_1=(2,1)$ and $T_2=(1,1)$, creating change scores of $G(\textnormal{\textit{plant}})=0.14$ and $B(\textnormal{\textit{plant}})=0$. The probabilities of each sense are relatively similar in the two corpora, which leads to a low graded change score and no binary change.

\begin{table}[htp]
  \tablecaption{Sample corpus split for the non-target lemma \textit{plant}. $T_1=(2,1)$, $T_2=(1,1)$, $G(w)=0.14$ and $B(w)=0$.}
  {
\begin{adjustbox}{width=1.0\linewidth}
\begin{tabular}{@{}p{6.8cm}p{6.8cm}@{}}\hline
 \centering $C_1$ & \centering $C_2$ \tabularnewline\hline
remove about half the branch from each \textbf{plant} leave only the strong with the largest bud \hfill \textbf{(s2)} & on the side toward the horizon the southern hemisphere it be spring \textbf{plant} are being teach to grow \hfill \textbf{(s2)}\\
the pilot \textbf{plant} was equip with a 3 hp turbine aerator figure 2 \hfill \textbf{(s1)} & can you share medical facility and staff with neighboring \textbf{plant} \hfill \textbf{(s1)}	\\
this reduce the number of expensive \textbf{plant} shutdown and startup \hfill \textbf{(s1)} & \\\hline
\end{tabular}\label{tab:split2}
\end{adjustbox}}
\end{table}%

\subsection{Testsets}

With the corpus split and the extracted change scores we have a large amount of evaluation data available. However, the change scores are subject to noise through non-annotated data. That is, non-annotated uses of words distort the sense frequency distributions on which the change scores are based. In order to minimize this noise we disregard each lemma $w$ that has a relative frequency error $RE(w)\geq 0.5$, where
\begin{equation*}
RE(w) = \frac{\#(w)-\#\textnormal{\textit{annotated}}(w)}{\#\textnormal{\textit{annotated}}(w)}
\end{equation*}
with $\#(w)$ being $w$'s corpus frequency and $\#\textnormal{\textit{annotated}}(w)$ the number of $w$'s annotated uses. Hence, we allow at most a number of half of $w$'s annotated uses to be added to these for $w$ to be part of the testset. We additionally disregard any lemma with a lower frequency than 50 in either of $C_1$, $C_2$. This results in a testset containing 148 lemmas with different change scores. All the data is publicly available and can be used for LSC detection evaluation.\footnote{Find the data under: \url{https://www.ims.uni-stuttgart.de/data/lsc-simul}.}

\subsection{Discussion}

The corpus splitting process described in Section \ref{sec:algorithm} controls the degree of change introduced for a particular lemma. However, this process is not built on a particular theoretical model of LSC, i.e., a model of how the underlying sense probability distributions should change to be similar to true LSC. This also determines how much variables such as polysemy and frequency will correlate with simulated LSC in the resulting dataset. The way in which we chose to split the corpus implicitly introduces higher rates of change for more polysemous words, i.e., simulated LSC correlates with polysemy. Similarly, it introduces specific frequency patterns for strongly changing words, i.e., simulated LSC correlates with frequency change. Whether and to which degree this holds for true LSC is still debated, but it is clear that these variables strongly bias model predictions \shortcite{Hellrich16p2785,dubossarsky2017}. Thus, to make sure that model performances on our dataset do not stem from model biases towards these variables we recommend to report a polysemy and a frequency baseline. Only performances above these baselines can be safely attributed not to stem from model biases.

\section{Model Evaluation}

We give a short example of how to evaluate LSC detection models on our dataset. We train all vector space models with all alignment techniques from \shortciteA{Schlechtwegetal19} on $C_1$ and $C_2$ and apply two similarity measures (CD, LND) to the resulting representations to create change score predictions.\footnote{Find implementations at \url{https://github.com/Garrafao/LSCDetection}. Because of the very small corpus size we choose a large window size of $n=10$ for all models, experiment with low dimensionalities $d=\{30,100\}$ for SVD, RI and SGNS and train all SGNS with 30 epochs. We set $k=5$ and $t=none$. The rest of parameters is set as in \shortciteA{Schlechtwegetal19}.} Then we use Spearman's $\rho$ to compare the resulting rankings against the graded change scores and Average Precision (AP) to compare them against the binary change scores. The results are presented in Table \ref{tab:results}.

\begin{table}[htp]
	\center
	\tablecaption{Best and mean $\rho$ (Graded) and AP (Binary) scores across similarity measures (SIM). Scores are averaged over five iterations for models with a random component. The column `model' gives the model with the best score. SGNS = Skip-Gram with Negative Sampling, CD = Cosine Distance, LND = Local Neighborhood Distance, SVD = Singular Value Decomposition, OP = Orthogonal Procrustes, WI = Word Injection, POLY = Polysemy Baseline, FREQ = Normalized Frequency Difference (NFD) Baseline, RAND = Approximate Random Baseline for Binary Classification.}{
	\begin{adjustbox}{width=1.0\linewidth}
	\begin{tabular}{ c c c c c c c c }
		\hline
	\multirow{2}{*}{\textbf{Dataset}}&\multirow{2}{*}{\textbf{Measure}} &\multicolumn{3}{c}{\textbf{Graded}} & \multicolumn{3}{c}{\textbf{Binary}}  \\	
	& &\textit{mean}&\textit{best} &\textit{model} &\textit{mean}&\textit{best} & \textit{model} \\
		\hline															
\multirow{4}{*}{\textbf{SEMCOR}} & SIM & 0.159 & \textbf{0.451} & SGNS+OP+CD & \textbf{0.182} & \textbf{0.376} & SVD+WI+LND \\\hdashline
		 & POLY & \textbf{0.349} & 0.349 & - & 0.151 & 0.151 & - \\
		 & FREQ & 0.120 & 0.120 & - & 0.110 & 0.110 & - \\
		 & RAND & - & - & - & 0.081 & 0.081 & - \\
		\hline               
	\end{tabular}
	\end{adjustbox}
\label{tab:results}}
\end{table}

Generally, models show rather weak performances on the testset. The performances for graded change are considerably lower than in \shortciteA{Schlechtwegetal19}, which may be attributed to the much smaller corpus sizes and the resulting noise. As expected, the frequency and polysemy baselines show positive correlations with change scores. On average the models outperform the frequency baseline for graded and binary change, while the polysemy baseline is only outperformed for binary change. However, the best models always outperform both baselines. Thus, we can conclude that a range of models measure more than just polysemy or frequency change.

The best models are SGNS and SVD with OP and WI (see Table \ref{tab:results}) as alignments. This is similar to previous results in that SGNS+OP+CD has outperformed other models and SVD showed generally high performance \shortcite{Schlechtwegetal19}. The comparably high performance of WI alignment may be attributed to its strong noise-reducing effect on our small and thus noisy training corpora \shortcite{Dubossarskyetal19}. A surprising observation is the performance of LND, as in the experiments of \citeauthor{Schlechtwegetal19} CD has constantly outperformed LND. This may be related to the difference between binary and graded change, as \citeauthor{Schlechtwegetal19} only evaluated on graded change.

\section{Conclusion}
\label{sec:conclusion}

We simulated lexical semantic change from synchronic sense-annotated data, introduced the first large-scale, synthetic gold standard for LSC detection and showed how to use it for evaluation. As part of our novel procedure, we provided quantitative definitions of various notions of LSC which implicitly underlie previous work; we thus provided a theoretical basis for artificial and empirical LSC detection evaluation. In the future, we will create further gold standards by exploiting sense-annotated data across languages and use our suggested LSC notions for the simulation of pseudo-change. We will also use the data to evaluate diachronic contextualized embeddings \shortcite{Giulianelli19,Hu19}.

The simulation procedure we proposed may also have applications in cognitive research on language evolution \shortcite{Karjus18,NOLLE201893,Tinits17} or more dialogue-oriented studies on meaning change \shortcite{PLEYER2017}, where it may be used to simulate the semantic development of words over generations or conversations. Similarly, different types of annotated data may be used to simulate specific types of LSC as e.g. literal and non-literal usages of words \shortcite{Koeper16}, metaphoric uses \shortcite{Koeper17} or concrete and abstract uses \shortcite{Naumann18}.

\section*{Acknowledgments}

The first author was supported by the Konrad Adenauer Foundation and the CRETA center funded by the German Ministry for Education and Research (BMBF).

\bibliographystyle{apacite}
\bibliography{200109-lsc-simulation}

\end{document}